%% file: neurips_2024.tex
\pdfoutput=1

\documentclass{article}
\usepackage{enumitem}

\usepackage{wrapfig}
\usepackage{amsmath}
\usepackage{amssymb}
\usepackage{algorithm}
\usepackage{algpseudocode}
\usepackage{tree-dvips}
\usepackage{qtree}
\usepackage{xcolor}
\definecolor{keywordcolor}{rgb}{0.7, 0.1, 0.1}   
\definecolor{tacticcolor}{rgb}{0.0, 0.1, 0.6}    
\definecolor{commentcolor}{rgb}{0.4, 0.4, 0.4}   
\definecolor{symbolcolor}{rgb}{0.0, 0.1, 0.6}    
\definecolor{sortcolor}{rgb}{0.1, 0.5, 0.1}      
\definecolor{attributecolor}{rgb}{0.7, 0.1, 0.1} 
\usepackage{caption}
\usepackage{subcaption}
\usepackage{multirow}
\usepackage{graphicx}
\usepackage{caption}
\usepackage{subcaption}
\usepackage{graphicx} 
\usepackage{caption}
\usepackage{subcaption} 
\usepackage{listings}

\usepackage[authoryear]{natbib}

\usepackage[utf8]{inputenc} 
\usepackage[T1]{fontenc}    
\usepackage{hyperref}       
\usepackage{url}            
\usepackage{booktabs}       
\usepackage{amsfonts}       
\usepackage{nicefrac}       
\usepackage{microtype}      
\usepackage{xcolor}         
\usepackage{graphicx}
\usepackage{algpseudocode}
\usepackage{algorithm}

\usepackage{amsmath}

    \usepackage[preprint]{neurips_2024}

\usepackage[utf8]{inputenc} 
\usepackage[T1]{fontenc}    
\usepackage{url}            
\usepackage{booktabs}       
\usepackage{amsfonts}       
\usepackage{nicefrac}       
\usepackage{microtype}      
\usepackage{graphicx}

\definecolor{citecolor}{HTML}{2980b9}
\definecolor{linkcolor}{HTML}{c0392b}
\usepackage{listings}

\newcounter{example}

\title{DeepSeek-Prover: Advancing Theorem Proving in LLMs  through Large-Scale Synthetic Data}

\author{Huajian Xin\textsuperscript{\rm 1,2}\quad  
Daya Guo\textsuperscript{\rm 1}\quad 
Zhihong Shao\textsuperscript{\rm 1}\quad 
\textbf{Z.Z. Ren\textsuperscript{\rm 1}\quad
Qihao Zhu\textsuperscript{\rm 1}\quad 
Bo Liu\textsuperscript{\rm 1}}\quad 
\\\textbf{Chong Ruan\textsuperscript{\rm 1}\quad 
Wenda Li\textsuperscript{\rm 3}\quad 
Xiaodan Liang\textsuperscript{\rm 2,4}\thanks{Corresponding author.}}\\
 $^1$DeepSeek\quad 
 $^2$Sun Yat-sen University\quad 
 $^3$University of Edinburgh\quad
 $^4$MBZUAI\quad  \\
\small\{xinhj, guoday, zhihongshao, rzz, zhuqh, chong.ruan\}@deepseek.com, \\
benjaminliu.eecs@gmail.com, wli8@ed.ac.uk, xdliang328@gmail.com
 }

\begin{document}

\maketitle

\begin{abstract}
Proof assistants like Lean have revolutionized mathematical proof verification, ensuring high accuracy and reliability. Although large language models (LLMs) show promise in mathematical reasoning, their advancement in formal theorem proving is hindered by a lack of training data. To address this issue, we introduce an approach to generate extensive Lean 4 proof data derived from high-school and undergraduate-level mathematical competition problems. This approach involves translating natural language problems into formal statements, filtering out low-quality statements, and generating proofs to create synthetic data. After fine-tuning the DeepSeekMath 7B model on this synthetic dataset, which comprises 8 million formal statements with proofs, our model achieved whole-proof generation accuracies of 46.3\% with 64 samples and 52\% cumulatively on the Lean 4 miniF2F test, surpassing the baseline GPT-4 at 23.0\% with 64 samples and a tree search reinforcement learning method at 41.0\%. Additionally, our model successfully proved 5 out of 148 problems in the Lean 4 Formalized International Mathematical Olympiad (FIMO) benchmark, while GPT-4 failed to prove any. These results demonstrate the potential of leveraging large-scale synthetic data to enhance theorem-proving capabilities in LLMs. Both the synthetic dataset and the model will be made available to facilitate further research in this promising field.
\end{abstract}

\section{Introduction}
In modern mathematics, the increasing complexity of proofs presents substantial challenges for peer review. This complexity has led to the acceptance of erroneous proofs, with critical flaws often detected only after considerable time. To address these issues, formal mathematical languages such as Lean \citep{de2015lean, moura2021lean}, Isabelle \citep{paulson_isabelle_1994}, and Coq \citep{coq} have been developed. These languages enable the creation of computer-verifiable proofs \citep{avigad2023mathematics}. However, crafting formal proofs demands significant effort, specialized expertise, and poses challenges even for seasoned mathematicians. Consequently, the significance of automated theorem proving is on the rise \citep{Shulman}.

To reduce the effort involved in writing formal mathematical proofs, several approaches \citep{polu2020generative, jiang2021lisa, han2021proof, polu2022formal, lample2022hypertree, jiang2022thor, yang2024leandojo} have been developed, primarily focusing on search algorithms that explore potential solutions for proposed theorems. However, these methods often struggle with the vast search spaces required for complex theorems, rendering them ineffective for more intricate proofs \citep{loos2017deep}. Recently, advances in large language models (LLMs) have introduced a novel strategy, utilizing pre-trained models to guide the search process. Although these new methods \citep{jiang2022draft, zhao2023decomposing, xin2023lego} represent significant improvements, they still fall short of practical applicability due to the lack of parallel corpus. Unlike conventional programming languages such as Python or Java, formal proof languages are used by relatively few mathematicians, resulting in limited datasets. Recent advances in autoformalization \citep{wu2022autoformalization} allow more aligned data to be synthesized to train LLM-based automated theorem provers. Nevertheless, the resulting dataset remains too small to fully unleash the capabilities of LLMs.

To address this issue, we propose a method for generating extensive Lean 4 proof data from informal mathematical problems. Our approach translates high-school and undergraduate-level mathematical competition problems into formal statements. We then automate proof generation using a large language model (LLM) and verify the correctness of these proofs within the Lean 4 environment. The primary challenge of this method is to ensure both the scale and quality of the synthetic data.

\textbf{Quality Assurance:} We enhance the quality of generated proofs through a multi-step process. First, we filter out simple statements using a quality scoring model and exclude invalid statements via a hypothesis rejection strategy. Our novel iterative framework then improves proof quality by initially generating synthetic statements from informal math problems using an under-trained LLM fine-tuned on limited data. These statements are used to generate corresponding proofs, which are validated for correctness using a Lean 4 verifier. The correct theorem-proof pairs are subsequently used to further train the initial model. Through several iterations, the model trained on large-scale synthetic data becomes significantly more powerful than the originally under-trained LLMs, resulting in higher-quality theorem-proof pairs.

\textbf{Scale Assurance:} To accelerate the proof generation process, our method addresses the challenge of the large search space for proofs. A significant cause of delays is the generation of unprovable statements that continue to be processed until they reach the time limit. To mitigate this, we propose proving negated statements in parallel. Once either the original statement or its negation is proved, the entire proving process is terminated.

We assess the effectiveness of our method on Lean 4 theorem proving using 488 problems from miniF2F \citep{zheng2021minif2f} and 148 problems from the FIMO benchmarks \citep{liu2023fimo}. We utilize DeepSeekMath 7B \citep{shao2024deepseekmath}, a state-of-the-art mathematical model, as our base. The results show that our iteratively trained model performs strongly, achieving 46.3\% accuracy in whole-proof generation on the miniF2F-test benchmark with 64 samples, surpassing GPT-4 \citep{achiam2023gpt} at 23.0\% and a reinforcement learning method at 41.0\%. Additionally, our approach solved 4 out of 148 problems in the FIMO benchmark with 100 samples, while GPT-4 solved none, and our approach solved 5 with 4096 samples. Ablation experiments indicate that the model progressively solves more problems in miniF2F with each iteration. In summary, our paper makes the following contributions:
\begin{itemize}
\item We introduce an iterative method to synthesize 8 million formal statements, each accompanied by a formal proof, from informal math problems. Experimental results demonstrate that this method significantly enhances both the scalability and quality of synthetic data.
\item Our model, trained on this synthetic dataset, achieves state-of-the-art performance on benchmarks, with whole-proof generation accuracies of 46.3\% using 64 samples and 52\% cumulatively on the Lean 4 miniF2F test. This surpasses the baseline GPT-4 at 23.0\% with 64 samples and a tree search reinforcement learning method at 41.0\%. Additionally, our model successfully proved 5 out of 148 problems in the Lean 4 Formalized International Mathematical Olympiad (FIMO) benchmark, while GPT-4 failed to prove any.
\item We contribute to the mathematical and AI communities by creating and open-sourcing a large dataset of high-quality formal mathematical proofs, thereby fostering further research and development in automated theorem proving.
\end{itemize}

\section{Background and Related Works}
\label{sec:backgroundandrelatedworks}
Automated theorem proving has been a significant area of interest in artificial intelligence research since its inception  \citep{bibel2013automated}. Initial efforts were directed at simpler logical frameworks, which led to the development of highly efficient first-order theorem provers like E   \citep{schulz2002brainiac} and Vampire   \citep{kovacs2013first}. Nonetheless, these tools often fall short in handling complex theorems commonly found in modern proof assistants such as Lean   \citep{de2015lean}, Isabelle   \citep{paulson_isabelle_1994}, and Coq   \citep{coq}. The advent of recent deep learning models and model-guided search techniques has reinvigorated the field   \citep{bansal2019holist}. This modern approach has not only enhanced the capabilities of ATP systems but also expanded their applicability in solving more intricate mathematical problems.


{\bf ATP with Neural Models.} With the development of deep learning, several approaches have been proposed to combine neural models with ATP  \citep{loos2017deep}. A series of ATP approaches adopts tree search algorithms guided by neural models  \citep{polu2020generative, han2021proof, polu2022formal, jiang2022thor, yang2024leandojo}. These approaches primarily utilize reinforcement learning techniques to enhance the accuracy of the model  \citep{kaliszyk2018reinforcement, crouse2021deep, wu2021tacticzero, lample2022hypertree}. Since the search space is significantly large, the searching process consumes considerable time and computing resources.

Another series of ATP approaches harnesses the power of large language models. These approaches typically involve language models that are fine-tuned with open-source proof data and interact with verifiers via a state-action transition program  \citep{polu2020generative, jiang2021lisa, han2021proof, polu2022formal, lample2022hypertree, jiang2022thor, yang2024leandojo}. This process iteratively generates proof steps and verifies their correctness with formal verifiers. It then generates the next proof steps based on the proof states returned by the formal verifiers. Although these approaches achieve high performance, they are computationally intensive. To enhance efficiency, recent researches leverage language models to generate complete formal proofs directly  \citep{first2023baldur, jiang2022draft, zhao2023decomposing, xin2023lego}, thus bypassing the iterative interaction during proof generation.

{\bf Autoformalization for Formal Mathematics.} 
Due to the limited availability of formal corpora for training, the performance of current large language models (LLMs) is also constrained. Thus, some approaches propose autoformalization  \citep{wu2022autoformalization, jiang2022draft}, which involves converting natural language descriptions into formal statements that can be verified by proof assistants. Several studies have generated synthetic datasets of formal proofs using rule-based transformations of existing theorems \citep{wu2020int, wang2020learning, xiong2023trigo}. While effective, these methods are constrained by their reliance on predefined rules and lack flexibility for broader applications. Recent methodologies adopt large language models to translating natural language problems into formal statements \citep{huang2024mustard}. However, these datasets remain smaller than needed and are limited to small mathematical benchmarks, leading to only minor improvements in training outcomes for language models. In this paper, we aim to synthesise formal proofs via autoformalization at a much larger scale to boost the performance of a neural prover.

\input{docs/method}

\input{docs/experiments}

\section{Conclusion}
\label{sec:conclusion}
In this paper, we presented a method to generate extensive synthetic proof data from high-school and undergraduate-level mathematical competition problems. By translating natural language problems into formal statements, filtering out low-quality ones, and using iterative proof generation, we created 8 million proof data points and significantly improved the DeepSeekMath 7B model's performance in ATP when trained on this synthetic data. Our model outperforms GPT-4 and other methods on benchmarks like miniF2F and FIMO. By open-sourcing our dataset and model, we aim to advance research in automated theorem proving and enhance the capabilities of large language models in formal mathematical reasoning. Currently, our work mainly focuses on algebra and number theory at the middle school and undergraduate levels. In future work, we will aim to expand the diversity of mathematical problems addressed, enhancing the general applicability of our methods in ATP.

\section*{Broader Impact}
\label{sec:broaderimpact}
The research presented in this paper has the potential to significantly advance automated theorem proving by leveraging large-scale synthetic proof data generated from informal mathematical problems. This remarkable advancement can enhance the capabilities of large language models in formal theorem proving, contributing to more reliable mathematical proof verification and providing valuable educational resources for students and researchers. By directly releasing the code, model, and data, we aim to ensure the responsible use of our work, fostering further innovation and maintaining high standards of data privacy and intellectual property compliance.

\bibliographystyle{abbrvnat}

\bibliography{biblio}

\input{docs/appendix}

\end{document}

%% file: docs/method.tex
\section{Approach}
\label{sec:approach}
\begin{figure}
   \centering
   \includegraphics[width=\textwidth]{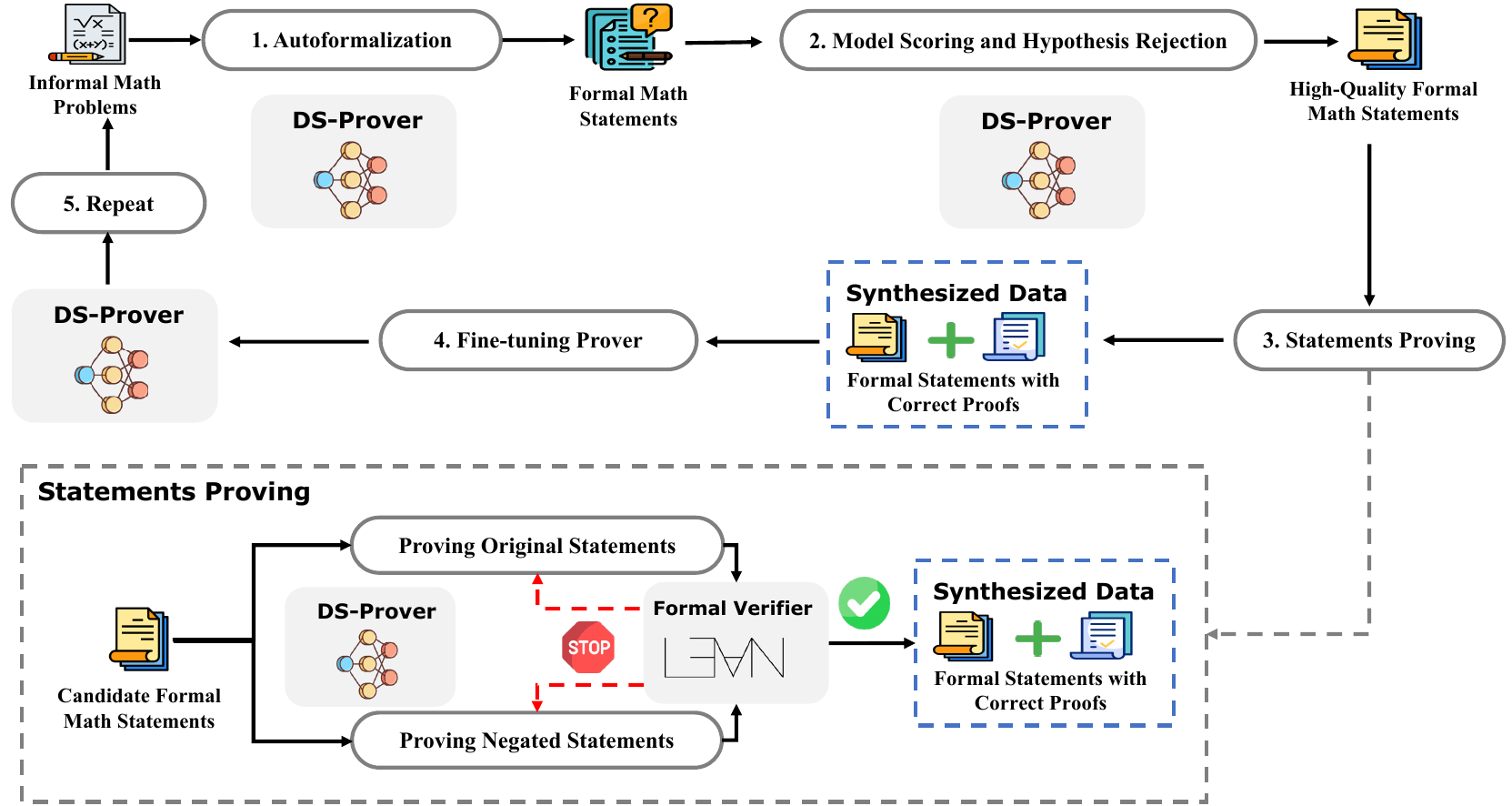}
  \caption{An overview of our approach.}
  \label{fig:overview}
\end{figure}
In this section, we introduce our approach, which consists of four key processes as depicted in Figure \ref{fig:overview}. The initial phase concentrates on generating formal mathematical statements from a broad collection of informal math problems, necessitating further proof. Next, the autoformalized statements are filtered through model scoring and hypothesis rejection methods to select high-quality statements. These statements are then proved by a model called DeepSeek-Prover, with their correctness verified by the formal verifier called Lean 4\footnote{$\mathtt{leanprover/lean4:v4.7.0-rc2}$}, yielding validated formal statements and proofs. These data serve as synthetic data for fine-tuning the DeepSeek-Prover. After enhancing DeepSeek-Prover, we repeat the entire previously described process. This cycle continues until the improvements in DeepSeek-Prover become marginal. Notably, to enhance proof efficiency, we prove concurrently both the original statements and their negations. This method has the advantage of swiftly discarding the original statement when it is invalid by proving its negation. The details of each phase will be described in the subsequent sections.

\subsection{Autoformalization}
The generation of formal proof data fundamentally relies on the availability of a substantial corpus of formal statements. In practice, however, amassing a large collection of manually crafted formal statements is challenging. Fortunately, the internet is replete with math-related problems expressed in natural language. By autoformalizing these informal mathematical problems, we can generate a vast repository of formal statements.

We have observed that problems with explicit conditions and well-defined goals are typically easier to formalize compared to advanced mathematical topics that necessitate intricate definitions and constructions. Consequently, this paper primarily examines high school and undergraduate-level competition problems, with a particular emphasis on algebra and number theory, and to a lesser extent, combinatorics, geometry, and statistics. Despite their apparent simplicity, these problems often involve complex solution techniques, making them excellent candidates for constructing proof data to improve theorem-proving capabilities in Large Language Models (LLMs). To compile our dataset, we employed web scraping and careful data cleaning techniques to extract problems from online resources featuring high school and undergraduate exercises, exams, and competitions, resulting in a dataset of 869,659 high-quality natural language math problems.

Specifically, we initialized the DeepSeek-Prover using the DeepSeekMath-Base 7B model \citep{shao2024deepseekmath}. Initially, the model struggled to convert informal math problems into formal statements. To address this, we fine-tuned the DeepSeek-Prover model using the MMA dataset \citep{jiang2023multilingual}, which comprises formal statements from Lean 4's mathlib\footnote{The specific mathlib commit used is $\mathtt{64528268b3c2cf578639bc479828882a9ecd3a82}$.} that were back-translated into natural language problem descriptions by GPT-4. We then instructed the model to translate these natural language problems into formal statements in Lean 4 using a structured approach.

\textbf{Prompt}:\\
\begin{minipage}[t!]{0.98\textwidth}
\centering
\fbox{%
  \parbox{\textwidth}{
    \texttt{Mathematical Problem in Natural Language:\\
    \{\$informal\_statement\_with\_answers\}\\
    Translate the problem to Lean 4 (only the core declaration): \\
    ```lean4}}
}
\end{minipage}

\textbf{Response}:\\
\begin{minipage}[t!]{0.98\textwidth}
\centering
\fbox{%
  \parbox{\textwidth}{
    \texttt{\{\$formal\_statement\}\\
    ```
    }
  }
}
\end{minipage}

\subsection{Quality Filtering}
The quality of the autoformalized statements was found to be suboptimal due to two main issues. Firstly, many formal statements were overly simplistic. To address this, we developed scoring criteria and provided examples from miniF2F-valid as few-shot examples to guide the DeepSeek-Prover model in evaluating the content and quality of these statements using a chain-of-thought approach. Manual review of these scores confirmed that the model's evaluations closely matched human intuition and expectations. Specifically, the model was instructed (see Appendix \ref{self-evalutation-prompt} for the detailed prompt) to classify the quality of each formal statement into categories: "excellent," "good," "above average," "fair," or "poor." Statements rated as "fair" or "poor" were subsequently excluded.

The second issue pertains to formal statements that, although provable, are based on inconsistent hypotheses leading to vacuous conclusions, rendering the conclusions meaningless in mathematics. For example, consider the following model-generated statement:

\begin{lstlisting}
example (θ : ℝ) (h₀ : ∀ z : ℂ, z ^ 2 = -1 ∧ z ^ 3 = -1 ∧ z ^ 6 = 1) (h₁ : Real.tan θ = 2 * Real.sqrt 3) : θ = 5 * Real.pi / 3
\end{lstlisting}

Here, the hypothesis $z ^ 2 = -1 \wedge z ^ 3 = -1 \wedge z ^ 6 = 1$ for all complex numbers is clearly false, making any derived conclusions meaningless. To eliminate such cases from our dataset, we implemented a hypothesis rejection method. This involves using the DeepSeek-Prover model to attempt proving the formal statement with 'False' as the conclusion. A successful proof indicates an invalid hypothesis, prompting exclusion of the statement. An example is shown below:

\begin{lstlisting}
example (θ : ℝ) (h₀ : ∀ z : ℂ, z ^ 2 = -1 ∧ z ^ 3 = -1 ∧ z ^ 6 = 1) (h₁ : Real.tan θ = 2 * Real.sqrt 3) : False := by
  simpa using h₀ 1
\end{lstlisting}

By applying this dual strategy of model scoring and hypothesis rejection, we curated a refined set of 712,073 high-quality formal statements, providing a robust foundation for further proof synthesis.

\subsection{Statement Proving}

After creating a substantial corpus of high-quality formal statements, we employed the model to search for proofs of these statements. Traditionally, language models have been used predominantly in a brute-force manner to prove theorems—repeatedly attempting until a valid proof is found or computational resources are exhausted. This approach is inefficient for our purposes. Typically, language models are applied to human-curated formal statements that are carefully crafted and generally true and provable; however, in our task of proving autoformalized statements, many of the statements produced by the model may be incorrect. Indeed, it is unreasonable to expect the model to validate a false proposition within any reliable proof system. This issue becomes more pronounced during large-scale autoformalization, where we observed that at least 20\% of the formal statements generated by our model, even after quality filtering, were incorrect, leading to significant computational waste if addressed with brute force.

To minimize resource wastage on unprovable statements and improve the efficiency of the proof search process, we exploited the logical symmetry between a statement and its negation to accelerate proof synthesis. We implemented dual concurrent proof searches for each synthetic statement—one for the statement $\Gamma \vdash P$ and another for its negation $\Gamma \vdash \neg P$. The search terminates as soon as a valid proof is found for either, conclusively demonstrating the unprovability of the other. Each proof search stream attempts up to $k$ proofs unless a valid proof emerges sooner.

All validated proofs, whether they justify the original theorems or their negations, are then aggregated to further train the DeepSeek-Prover. Thus, this dual approach serves as a form of data augmentation, enriching the dataset with both propositions and their negations—even if the original propositions were not correctly formalized by the model.

\subsection{Iterative Enhancement}
Since the entire pipeline heavily relies on the DeepSeek-Prover, enhancing the model's performance after each iteration is crucial. To achieve this, we consistently fine-tune the model with newly generated data. The updated model is then utilized for subsequent autoformalization iterations. The key insight from this iterative process is that the model incrementally improves in strength and efficacy after each cycle of refinement and application. This iterative process continues until no further gains are observed. Consequently, the theorem-proof pairs generated by the model become increasingly higher in quality with each iteration. This method ensures that the DeepSeek-Prover consistently enhances its performance, ultimately producing superior theorem-proof pairs through continuous refinement.

%% file: docs/experiments.tex
\section{Experiments}
\label{sec:experiments}
\subsection{Experimental Setup}

DeepSeek-Prover is built upon DeepSeekMath-Base 7B model \citep{shao2024deepseekmath}, a decoder-only transformer  \citep{vaswani2017attention} pre-trained on a corpus comprising 120 billion math-related tokens. We fine-tuned this model using a global batch size of 512 and a constant learning rate of $1 \times 10^{-4}$, incorporating 6,000 warmup steps with synthetic data. DeepSeek-Prover's performance was evaluated against several baselines:

\begin{itemize}
    \item \textbf{GPT-3.5 and GPT-4} \citep{achiam2023gpt}, developed by OpenAI, are advanced generative AI models known for their effectiveness in diverse tasks, including code generation. Although not explicitly designed for theorem proving, their extensive scale and parameter count confer significant capabilities. In contrast, \textbf{DeepSeekMath} is a specialized model, explicitly pre-trained for mathematical content. We utilized both GPT-4 (specifically the GPT-4-turbo 0409 version) and DeepSeekMath to generate complete proofs for given theorems using a methodology similar to ours.

    \item \textbf{GPT-f} \citep{polu2020generative}, utilizing a GPT-2-inspired architecture \citep{radford2019language}, implements an iterative best-first search method to progressively generate and validate proof steps within a formal proof setting until a proof is either completed or resources are depleted. This methodology has been further advanced by \textbf{Proof Artifact Co-Training} \citep{han2021proof}, \textbf{ReProver} \citep{yang2024leandojo}, \textbf{Llemma} \citep{azerbayev2023llemma}, and \textbf{COPRA} \citep{thakur2023language}, which employ either specialized fine-tuned models or versatile general-purpose models such as GPT-3.5 and GPT-4 for the generation of proof steps.
\end{itemize}

\subsection{Main Results}

This study addresses complex mathematical problems in algebra and number theory. We evaluate the theorem-proving efficacy of our model using the miniF2F \citep{zheng2021minif2f} and FIMO \citep{liu2023fimo} benchmarks. The metric pass@k is employed to denote the scenario where at least one valid proof is discovered among the first k attempts generated by the model.


\begin{table*}[tbh]
\begin{center}
\caption{
Comparing with state-of-the-arts on the miniF2F dataset.
}
\label{tab:main_results} 
\small
\begin{tabular}{lcccc}
\toprule
    Method & Model size & Generation times & miniF2F-valid & miniF2F-test \\
    \midrule
    \multicolumn{5}{l}{\textit{Tree Search Methods}} \\
    \midrule
     COPRA  (GPT-3.5) \citep{thakur2023language} & - & $1\times 60$ & - & $9.0\%$ \\
     COPRA (GPT-4) \citep{thakur2023language}& - & $1\times 60$ & - & $26.6\%$ \\
    \multirow{2}{*}{Proof Artifact Co-Training   \citep{han2021proof}} & \multirow{2}{*}{837M} & $1\times 8\times 512$ & $23.9\%$ & $24.6\%$ \\
    &  & $8\times 8\times 512$ & $29.3\%$ & $29.2\%$ \\
     ReProver   \citep{yang2024leandojo} & 229M & $1\times 3751$ & - & $25.0\%$ \\
     Llemma   \citep{azerbayev2023llemma} & 7B & $1\times 3200$ & - & $26.2\%$ \\
     Llemma   \citep{azerbayev2023llemma} & 34B & $1\times 3200$ & - & $25.8\%$ \\
    \multirow{3}{*}{Curriculum Learning   \citep{polu2022formal}} & \multirow{3}{*}{837M} & $1\times 8\times 512$ & $33.6\%$ & $29.6\%$ \\
     &  & $8\times 8\times 512$ & $41.2\%$ & $34.5\%$ \\
     &  & $64\times 8\times 512$ & $47.3\%$ & $36.6\%$ \\
     \multirow{2}{*}{Hypertree Proof Search   \citep{lample2022hypertree}} & \multirow{2}{*}{600M} & cumulative & $58.6\%$ & - \\
     &  & $64\times 5000$ & - & $41.0\%$ \\
     \midrule
    \multicolumn{5}{l}{\textit{Whole-Proof Generation Methods}} \\
    \midrule
    GPT-4-turbo 0409 & - & 64 & $25.4\%$ & $23.0\%$ \\
    DeepSeekMath-Base \citep{shao2024deepseekmath} & 7B & 128 & $25.4\%$ & $27.5\%$ \\
    \multirow{6}{*}{DeepSeek-Prover} & \multirow{6}{*}{7B} & cumulative & \textbf{60.2\%} & \textbf{52.0\%} \\
     &  & $1$ (greedy) & - & $30.0\%$ \\
     &  & $64$ & - & $46.3\%$ \\
     &  & $128$ & - & $46.3\%$ \\
     &  & $8192$ & - & $48.8\%$ \\
     &  & $65536$ & - & $50.0\%$ \\
    \bottomrule
\end{tabular}
\end{center}

\end{table*}

\textbf{Results on MiniF2F.}
The miniF2F benchmark consists of 244 validation and 244 test problems, ranging from basic arithmetic to competition-level problems, e.g., problems from the American Invitational Mathematics Examination (AIME), the American Mathematics Competitions (AMC), and the International Mathematical Olympiad (IMO). We use the version of miniF2F in Lean 4, which was released by the LeanDojo project (\url{https://github.com/yangky11/miniF2F-lean4}).

Table \ref{tab:main_results} compares various state-of-the-art methods on the miniF2F dataset. DeepSeek-Prover outperforms all with cumulative scores of 60.2\% on miniF2F-valid and 52.0\% on miniF2F-test, significantly higher than other methods, including GPT-4 which scores 25.41\% and 22.95\%, respectively. Even the best tree search method, Hypertree Proof Search with a 600M model, achieves only up to 58.6\% on miniF2F-valid and 41.0\% on miniF2F-test. DeepSeek-Prover's scalability is evident as its performance improves with increased computational resources, rising from 30.03\% using a greedy approach  to 50.0\% at 65536 generation times, demonstrating its effectiveness in handling complex proof scenarios. Examples of proved theorems of MiniF2F can be found in Appendix \ref{minif2f_examples}.

\textbf{Results on FIMO.}
The FIMO benchmark comprises 149 formal problems which are sourced from the IMO shortlist translated into Lean 4.
Our method successfully proved 4 theorems with 100 attempts per theorem, whereas GPT-4 failed to prove any.
By increasing the number of attempts per theorem to 4,096, we successfully proved an additional theorem. Examples of proved theorems of FIMO can be found in Appendix \ref{fimo_examples}.

\subsection{Ablation Studies}

\subsubsection{The Effectiveness of Large-scale Autoformalization}
To demonstrate the effectiveness of large-scale autoformalization, we conducted a comparative analysis as shown in Table \ref{tab:ablation_data_source} between our autoformalized dataset and conventional datasets using expert iteration \citep{polu2020generative}. This iterative method entails generating formal proofs, fine-tune the model based on successful outcomes, and iterating this process until no additional enhancements are observed. The results indicate that models trained with our autoformalized data significantly outperform those trained solely with mathlib data.

\begin{table*}[tbh]

\begin{center}
\caption{
Improvement in pass rates for miniF2F at pass@128 in models trained on formal proofs, including those derived from human-authored theorems in Lean 4's mathlib and automatically formalized theorems.
}
\label{tab:ablation_data_source} 
\small
\begin{tabular}{lcccc}

\toprule
    Model & \#Tokens & miniF2F-valid & miniF2F-test \\
    \midrule
    - & - & $25.4\%$ & $27.5\%$ \\
    Mathlib & $0.238$B  & $30.3\%$ & $31.2\%$ \\
    Autoformalized Statements & $3.108$B  & $48.8\%$ & $42.6\%$ \\
    \bottomrule
\end{tabular}
\end{center}


\end{table*}

\subsubsection{The Effectiveness of Formal Statements Scoring}
To demonstrate the effectiveness of the model in filtering out low-quality statements, we fine-tuned the DeepSeekMath-Base model using an equal amount of high-score proof data and low-score proof data to verify the quality of the data, as shown in Table \ref{tab:score_pass_rate}. The table shows that the model trained on high-score proof data outperformed the model trained on low-score proof data by 4.5\%. This enhancement underscores the utility of the model in accurately scoring and effectively filtering out lower-quality statements.

\begin{table*}[tbh]
\begin{center}
\caption{Improvement in pass rates for miniF2F at pass@128 in models trained on differently scored proof data.}
\label{tab:score_pass_rate}
\small
\begin{tabular}{lcc}
\toprule
    Scored Class  & miniF2F-valid & miniF2F-test \\
    \midrule
     "excellent", "good" and "above average" & $48.8\%$ & $42.6\%$ \\
     "fair" and "poor" & $41.4\%$ & $38.1\%$ \\
    \bottomrule
\end{tabular}
\end{center}

\end{table*}


\subsubsection{The Effectiveness of Iterative Enhancement}
Table \ref{tab:ablation_data_iteration} demonstrates a distinct correlation between the number of iterations in data synthesis and enhanced performance in theorem proving. This evidence underscores the success of our iterative enhancement strategy in augmenting theorem-proving capabilities. Successive iterations not only refine the model's ability to handle complex proofs but also significantly increase the quality and quantity of the synthetic data produced.

\begin{table*}[tbh]
\begin{center}
\caption{
Improvement in pass rates for miniF2F at pass@128 in models across successive training iterations, facilitated by the incremental integration of synthesized data via autoformalization.
}
\label{tab:ablation_data_iteration} 
\small
\begin{tabular}{lcc}
        \toprule
            Model &  miniF2F-valid & miniF2F-test \\
            \midrule
            iteration 0 & $38.1\%$ & $34.0\%$ \\
            iteration 1 & $45.1\%$ & $39.3\%$ \\
            iteration 2 & $49.2\%$ & $41.4\%$ \\
            iteration 3 & $54.5\%$ & $45.1\%$ \\
            iteration 4 & $59.4\%$ & $46.3\%$ \\
            \bottomrule
        \end{tabular}
\end{center}

\end{table*}

\subsubsection{The Effectiveness of Scaling Synthetic Theorem Proving Data}
Our investigation into synthetic theorem proving data reveals a clear correlation between dataset size and model efficacy, as illustrated in Table \ref{tab:ablation_synthetic_data_size}. By examining subsets of the eight million generated proof data points, we observed that performance on the miniF2F benchmark improves proportionally to the exponential increase in dataset size. This pattern highlights the pivotal importance of large-scale datasets for boosting model proficiency in automatically formalizing natural language questions. These findings emphasize the significant potential and necessity of systematic data construction for progressing in the field of automated theorem proving.

\begin{table*}[tbh]
\begin{center}
\caption{
Improvement in pass rates for miniF2F at pass@128 in models trained with a larger fraction of synthesized data via autoformalization.
}
\label{tab:ablation_synthetic_data_size} 
\small
\begin{tabular}{lcc}
        \toprule
            Size & miniF2F-valid & miniF2F-test \\
            \midrule
             1,000 & $22.95\%$ & $24.18\%$ \\
             10,000 & $32.79\%$ & $31.97\%$ \\
             100,000 & $36.07\%$ & $37.7\%$ \\
             1,000,000 & $39.34\%$ & $38.11\%$ \\
             8,066,621 & $42.62\%$ & $40.16\%$ \\
            \bottomrule
        \end{tabular}
\end{center}

\end{table*}
\section{Case Studies}

This section presents two case studies to demonstrate the application of our methods in autoformalizing theorems. It showcases both successful proofs and the identification of inconsistencies during the Hypothesis Rejection stage.

\subsection{Autoformalized Theorem with Complete Proof}

{\bf Example a.} {\it Problem}: Prove that the determinant of the following matrix  is zero.
$$
\begin{bmatrix}
1 & \cos(a - b) & \cos(a) \\
\cos(a - b) & 1 & \cos(b) \\
\cos(a) & \cos(b) & 1
\end{bmatrix}
$$

{\it Autoformalized Theorem in Lean}:
\begin{lstlisting}
example (a b : ℝ) :
  Matrix.det ![![1, Real.cos (a - b), Real.cos a], ![Real.cos (a - b), 1, Real.cos b], ![Real.cos a, Real.cos b, 1]] = 0 
\end{lstlisting}

This approach effectively translates the algebraic expression of the matrix and its determinant into a formal language using Lean. The autoformalization captures the essence of the original mathematical statement by defining a specific $3\times 3$ matrix dependent on real numbers \(a\) and \(b\), and asserts that its determinant is zero. The formalization employs the \texttt{Matrix.det} function to compute the determinant, utilizing the \texttt{![...]} notation for lists of lists in Lean to represent the matrix rows.

\subsection{Autoformalization of Theorem with Inconsistent Hypotheses}

{\bf Example b.} {\it Problem}: Given a real number \( D \) and the condition that for non-zero real numbers \( a, b, c \), the determinant of the matrix \( \begin{bmatrix} a & b & c \\ 1 & 4 & 9 \\ 3 & 1 & 2 \end{bmatrix} \) equals \( D \), prove that \( D^2 = 154 \).

{\it Autoformalized Theorem in Lean}:
\begin{lstlisting}
example (D : ℝ) (h₀ : ∀ a b c : ℝ, a ≠ 0 ∧ b ≠ 0 ∧ c ≠ 0 →
  Matrix.det ![![a, b, c], ![1, 4, 9], ![3, 1, 2]] = D) : D ^ 2 = 154
\end{lstlisting}

The initial autoformalization incorrectly assumes that the condition \(D^2 = 154\) universally applies to all non-zero real numbers \(a\), \(b\), and \(c\). This assumption is not supported by the problem statement, which does not claim universal applicability. Instead, the formalization should aim to either identify specific values of \(a\), \(b\), and \(c\) that satisfy \(D^2 = 154\) or demonstrate that no such values exist.

The model successfully identifies this inconsistency and provides a counterexample to demonstrate the absurdity of the hypothesis:

\begin{lstlisting}
example (D : ℝ) (h₀ : ∀ a b c : ℝ, a ≠ 0 ∧ b ≠ 0 ∧ c ≠ 0 →
  Matrix.det ![![a, b, c], ![1, 4, 9], ![3, 1, 2]] = D) : False := by
  have h₁ := h₀ 1 2 3
  have h₂ := h₀ 1 4 9
  simp [Matrix.det_fin_three] at h₁ h₂
  linarith
\end{lstlisting}

A corrected version of the autoformalized theorem can be proposed as follows:

\begin{lstlisting}
example (a b c : ℝ) (h₀ : a ≠ 0 ∧ b ≠ 0 ∧ c ≠ 0) :
  let D := Matrix.det ![![a, b, c], ![1, 4, 9], ![3, 1, 2]];
  D ^ 2 = 154
\end{lstlisting}

These examples illustrate the model's capability to verify proofs and identify hypothesis inconsistencies effectively. Further details can be found in Appendix \ref{autoformalization_cases}.

%% file: docs/appendix.tex
\newpage
\appendix

\section{Appendix / supplemental material}

\subsection{Prompts}

Specifically, we use the following format for scoring for the quality of the formalized statements:

\textbf{Prompt}:\\
\begin{minipage}[t!]{0.98\textwidth}
\label{self-evalutation-prompt}
\centering
\fbox{%
  \parbox{\textwidth}{
    \texttt{To evaluate whether a formal Lean4 statement will be of interest to the community, consider the following criteria:\\\\
1. Relevance to Current Research: Does the statement address a problem or concept that is actively being researched in mathematics or related fields? Higher relevance scores indicate greater potential interest.\\\\
2. Complexity and Depth: Is the statement complex enough to challenge existing theories and methodologies, yet deep enough to provide significant insights or advancements? Complexity and depth showcase Lean4's capabilities and attract interest.\\\\
3. Interdisciplinary Potential: Does the statement offer opportunities for interdisciplinary research, connecting mathematics with other fields such as computer science, physics, or biology? Interdisciplinary projects often garner wide interest.\\\\
4. Community Needs and Gaps: Does the statement fill an identified need or gap within the Lean4 community or the broader mathematical community? Addressing these needs directly correlates with interest.\\\\
5. Innovativeness: How innovative is the statement? Does it propose new methods, concepts, or applications? Innovation drives interest and engagement.\\\\
Customize your evaluation for each problem accordingly, assessing it as `excellent`, `good`, `above average`, `fair` or `poor`.\\\\
You should respond in the following format for each statement:\\\\
```\\\\
Translate the code to natural language: (Detailed explanation of the informal statement, including any relevant background information, assumptions, and definitions.)\\
Analysis: (Provide a brief justification for each score, highlighting why the statement scored as it did across the criteria.)\\
Assessment: (Based on the criteria, rate the statement as `excellent', `good`, `above average`, `fair` or `poor`.)\\
```
}}
}
\end{minipage}

\subsection{Case Studies of Autoformalization}

\label{autoformalization_cases}








{\bf Example a.}
{\it Problem in Natural Language}: For a real number $a$ and a function $f$ defined on real numbers, where $f(x) = x^3 - ax - 1$, if for all $x$, $f(x) \leq 0$ implies $x$ is either less than $-1$ or greater than $1$, then $a$ must equal $3$.

{\it Autoformalized Theorems with Complete Proofs}:
\begin{lstlisting}
example (a : ℝ) (f : ℝ → ℝ) (h₀ : ∀ x, f x = x ^ 3 - a * x - 1) :
    (∀ x, f x ≤ 0 → x ∈ Set.Iio (-1) ∪ Set.Ioi 1) → a = 3 := by
  intro h₁  
  have h₂ := h₁ 0  
  simp [h₀] at h₂  
  have h₃ := h₁ 1  
  simp [h₀] at h₃  
  have h₄ := h₁ (-1)  
  simp [h₀] at h₄
  linarith
\end{lstlisting}

{\it Analysis:} This is a simple example to illustrate common cases in the autoformalization procedure. The model is required to recognize natural language descriptions in the problem statement and formalize them into proper mathematical operators. In this case, the statement "\textit{either less than $-1$ or greater than $1$}" describes a union of interval $(-\infty, -1)\cup(1,+\infty)$, which has been accurately formalized to \texttt{Set.Iio (-1) $\cup$ Set.Ioi 1}, where \texttt{Set.Iio (-1)} denotes an infinite-open interval ends at $-1$, and \texttt{Set.Ioi 1} denotes an open-infinite interval begins at $1$.

{\bf Example b.}
{\it Problem in Natural Language:} Define two functions, $ F $ and $ f $, where $ F(k) $ is the binomial coefficient of $(k + 2)$ choose 2, and $ f(k) $ is the binomial coefficient of $(k + 1)$ choose 1. For any natural number $ n $ greater than or equal to 2, there exists a natural number $ k $ such that $ f(k) $ equals $(n - 1)$ times $ n $ divided by 2. Furthermore, there exists a natural number $ m $ such that $ F(m) $ equals $(n - 1)$ times $ n $ divided by 2, implying $ m $ equals $ k $.

{\it Autoformalized Theorems with Complete Proofs:}
\begin{lstlisting}
example : 
  let F := fun k => Nat.choose (k + 2) 2;
  let f := fun k => Nat.choose (k + 1) 1;
  ∀ n : ℕ, 2 ≤ n → ∃ k : ℕ, f k = (n - 1) * n / 2 → (∃ m : ℕ, F m = (n - 1) * n / 2 → m = k) := by
  simp [Nat.choose, Nat.mul_sub_left_distrib, Nat.sub_sub, Nat.mul_one]
  aesop
\end{lstlisting}

{\it Analysis:} The formalization for this combinatorial problem involving binomial coefficients is semantically faithful and effective. The use of $\mathtt{Nat.choose}$ accurately represents the binomial coefficients, and the logical structure with existential quantifiers ($\exists$) and implications ($\rightarrow$) aligns well with the problem's requirements.

{\bf Example c.}
{\it Problem in Natural Language:} Given three 3-dimensional vectors $a$, $b$, and $c$, where the dot product of $a$ and $b$ equals $-3$, the dot product of $a$ and $c$ equals $4$, and the dot product of $b$ and $c$ equals $6$. Prove that the dot product of $b$ with $7c - 2a$ equals $48$.

{\it Autoformalized Theorems with Complete Proofs:}
\begin{lstlisting}
example (a b c : ℝ × ℝ × ℝ)
  (h₀ : a.1 * b.1 + a.2.1 * b.2.1 + a.2.2 * b.2.2 = -3)
  (h₁ : a.1 * c.1 + a.2.1 * c.2.1 + a.2.2 * c.2.2 = 4)
  (h₂ : b.1 * c.1 + b.2.1 * c.2.1 + b.2.2 * c.2.2 = 6)
  : b.1 * (7 * c.1 - 2 * a.1) + b.2.1 * (7 * c.2.1 - 2 * a.2.1) + b.2.2 * (7 * c.2.2 - 2 * a.2.2) = 48 := by
  linarith [h₀, h₁, h₂]
\end{lstlisting}

{\it Analysis:} This formalization result uses Cartesian product $\mathbb{R}\times\mathbb{R}\times\mathbb{R}$ to represent a 3-dimensional vector, where the components of a 3-dimensional Cartesian product \texttt{a} are denoted by \texttt{a.1}, \texttt{a.2.1}, and \texttt{a.2.2} in Lean. The notation of dot products is expanded to the sum of element-wise products by definition.

\subsection{Example Lean Proofs Generated by Our Model}

\subsubsection{Results on MiniF2F-Test Dataset}
\label{minif2f_examples}

{\bf Example a.}
{\it Problem in Natural Language:}
Show that for any real number $x$ and any natural number $n$, if $x > -1$, then $(1+nx)\leq (1+x)^n$.

{\it Formal Proof:}
\begin{lstlisting}
theorem induction_1pxpownlt1pnx (x : ℝ) (n : ℕ) (h₀ : -1 < x) (h₁ : 0 < n)
  : 1 + ↑n * x ≤ (1 + x) ^ (n : ℕ) := by
  induction' h₁ with k h₁ <;> simp_all [Nat.succ_eq_add_one, pow_add, mul_add, add_assoc, add_comm, add_left_comm]
  nlinarith [mul_self_nonneg (1 + x - 1)]
\end{lstlisting}

{\bf Example b.}
{\it Problem in Natural Language:}
Ms. Blackwell gives an exam to two classes. The mean of the scores of the students in the morning class is $84$, and the afternoon class's mean score is $70$. The ratio of the number of students in the morning class to the number of students in the afternoon class is $\frac{3}{4}$. What is the mean of the scores of all the students?
Show that it is 76.

{\it Formal Proof:}
\begin{lstlisting}
theorem amc12b_2021_p4 (m a : ℕ) (h₀ : 0 < m ∧ 0 < a)
  (h₁ : ↑m / ↑a = (3 : ℝ) / 4)
  : (84 * ↑m + 70 * ↑a) / (↑m + ↑a) = (76 : ℝ) := by
  have h₂ := h₀.1.ne'
  have h₃ := h₀.2.ne'
  field_simp at h₂ h₃ ⊢
  ring_nf
  norm_num
  rw [div_eq_inv_mul] at h₁
  field_simp at h₁
  linarith
\end{lstlisting}

{\bf Example c.}
{\it Problem in Natural Language:}
For how many positive integers $m$ does there exist at least one positive integer n such that $m \cdot n \le m + n$? Show that it is infinitely many.

{\it Formal Proof:}
\begin{lstlisting}
theorem amc12a_2002_p6 (n : ℕ) (h₀ : 0 < n)
  : ∃ m, m > n ∧ ∃ p, m * p ≤ m + p := by
  simp_all only [ge_iff_le, gt_iff_lt, mul_one, mul_add, mul_comm, mul_assoc, mul_left_comm]
  use n + 1
  constructor
  exact Nat.lt_succ_self n
  use 1
  ring_nf
  nlinarith
\end{lstlisting}

\subsubsection{Results on FIMO Dataset}
\label{fimo_examples}
{\bf Example a.}
{\it Problem in Natural Language:}
Determine all functions $f$ from the set of positive integers into the set of positive integers such that for all $x$ and $y$ there exists a non degenerated triangle with sides of lengths $x$, $f(y)$ and $f(y+f(x)-1)$.
The final answer is the identity function $f(x)=x$.

{\it Formal Proof:}
\begin{lstlisting}
theorem fimo_2009_algebra_p3
  (f : ℕ → ℕ)
  (h₀ : ∀ x y, ∃ (a b c : ℕ),
    a = x ∧
    b = f y ∧
    c = f (y + f x - 1) ∧
    a + b > c ∧
    a + c > b ∧
    b + c > a) :
  ∀ x, f x = x := by 
  have h₁ : f 0 = 0 := by
    specialize h₀ 0 0
    obtain ⟨_, _, _, h₁, h₂, h₃⟩ := h₀
    simp at *
    linarith
  intro x
  induction x with
  | zero => exact h₁
  | succ x ih =>
    specialize h₀ (x + 1) 0
    obtain ⟨_, _, _, h₁, h₂, h₃⟩ := h₀
    simp at *
    linarith
\end{lstlisting}

{\bf Example b.}
{\it Problem in Natural Language:}
Prove that for every positive integer $n$, there exists a fraction $\frac{a}{b}$ where $a$ and $b$ are integers satisfying $0<b \leqslant \sqrt{n}+1$ and $\sqrt{n} \leqslant \frac{a}{b} \leqslant \sqrt{n+1}$.

{\it Formal Proof:}
\begin{lstlisting}
theorem fimo_2016_algebra_p5_1
  (n : ℕ)
  (h₀ : 0 < n) :
  ∃ a b : ℕ, 0 < b ∧ b ≤ n.sqrt + 1 ∧ 
    n.sqrt ≤ a / b ∧ a / b ≤ (n + 1).sqrt := by
  use n.sqrt, 1
  constructor <;>
  simp [h₀, Nat.sqrt_le_sqrt] <;>
  norm_num <;>
  linarith [Nat.sqrt_le_sqrt n]
\end{lstlisting}

\subsection{Formal Verification Details}
\label{lean4_environment}
We verify the generated Lean 4 code with the following code as the prefix:

\begin{lstlisting}
import Mathlib.Algebra.Algebra.Basic
import Mathlib.Algebra.Order.Floor
import Mathlib.Algebra.Associated
import Mathlib.Algebra.BigOperators.Basic
import Mathlib.Algebra.BigOperators.Order
import Mathlib.Algebra.BigOperators.Pi
import Mathlib.Algebra.GeomSum
import Mathlib.Algebra.Group.Pi.Basic
import Mathlib.Algebra.Group.Commute.Basic
import Mathlib.Algebra.GroupPower.Basic
import Mathlib.Algebra.GroupPower.Identities
import Mathlib.Algebra.Order.Floor
import Mathlib.Algebra.QuadraticDiscriminant
import Mathlib.Algebra.Ring.Basic
import Mathlib.Analysis.Asymptotics.AsymptoticEquivalent
import Mathlib.Analysis.NormedSpace.Basic
import Mathlib.Analysis.SpecialFunctions.Log.Basic
import Mathlib.Analysis.SpecialFunctions.Log.Base
import Mathlib.Combinatorics.SimpleGraph.Basic
import Mathlib.Data.Complex.Basic
import Mathlib.Data.Complex.Exponential
import Mathlib.Data.Finset.Basic
import Mathlib.Data.Fintype.Card
import Mathlib.Data.Int.Basic
import Mathlib.Data.Int.GCD
import Mathlib.Data.Int.ModEq
import Mathlib.Data.Int.Parity
import Mathlib.Data.List.Intervals
import Mathlib.Data.List.Palindrome
import Mathlib.Data.Multiset.Basic
import Mathlib.Data.Nat.Basic
import Mathlib.Data.Nat.Choose.Basic
import Mathlib.Data.Nat.Digits
import Mathlib.Data.Nat.Factorial.Basic
import Mathlib.Data.Nat.ModEq
import Mathlib.Data.Nat.Multiplicity
import Mathlib.Data.Nat.Parity
import Mathlib.Data.Nat.Prime
import Mathlib.Data.PNat.Basic
import Mathlib.Data.PNat.Prime
import Mathlib.Data.Polynomial.Basic
import Mathlib.Data.Polynomial.Eval
import Mathlib.Data.Real.Basic
import Mathlib.Data.Real.Irrational
import Mathlib.Data.Real.NNReal
import Mathlib.Data.Real.Sqrt
import Mathlib.Data.Set.Finite
import Mathlib.Data.Sym.Sym2
import Mathlib.Data.ZMod.Basic
import Mathlib.Dynamics.FixedPoints.Basic
import Mathlib.LinearAlgebra.AffineSpace.AffineMap
import Mathlib.LinearAlgebra.AffineSpace.Independent
import Mathlib.LinearAlgebra.AffineSpace.Ordered
import Mathlib.LinearAlgebra.FiniteDimensional
import Mathlib.Logic.Equiv.Basic
import Mathlib.Order.Filter.Basic
import Mathlib.Order.LocallyFinite
import Mathlib.Order.WellFounded
import Mathlib.Topology.Basic
import Mathlib.Topology.Instances.NNReal
import Aesop

set_option maxHeartbeats 0
set_option trace.aesop true
set_option trace.aesop.proof true

open Nat Real Rat BigOperators
\end{lstlisting}

%% file: neurips_2024.bbl
\begin{thebibliography}{39}
\providecommand{\natexlab}[1]{#1}
\providecommand{\url}[1]{\texttt{#1}}
\expandafter\ifx\csname urlstyle\endcsname\relax
  \providecommand{\doi}[1]{doi: #1}\else
  \providecommand{\doi}{doi: \begingroup \urlstyle{rm}\Url}\fi

\bibitem[Achiam et~al.(2023)Achiam, Adler, Agarwal, Ahmad, Akkaya, Aleman, Almeida, Altenschmidt, Altman, Anadkat, et~al.]{achiam2023gpt}
J.~Achiam, S.~Adler, S.~Agarwal, L.~Ahmad, I.~Akkaya, F.~L. Aleman, D.~Almeida, J.~Altenschmidt, S.~Altman, S.~Anadkat, et~al.
\newblock Gpt-4 technical report.
\newblock \emph{arXiv preprint arXiv:2303.08774}, 2023.

\bibitem[Avigad(2023)]{avigad2023mathematics}
J.~Avigad.
\newblock Mathematics and the formal turn, 2023.

\bibitem[Azerbayev et~al.(2023)Azerbayev, Schoelkopf, Paster, Santos, McAleer, Jiang, Deng, Biderman, and Welleck]{azerbayev2023llemma}
Z.~Azerbayev, H.~Schoelkopf, K.~Paster, M.~D. Santos, S.~McAleer, A.~Q. Jiang, J.~Deng, S.~Biderman, and S.~Welleck.
\newblock Llemma: An open language model for mathematics.
\newblock \emph{arXiv preprint arXiv:2310.10631}, 2023.

\bibitem[Bansal et~al.(2019)Bansal, Loos, Rabe, Szegedy, and Wilcox]{bansal2019holist}
K.~Bansal, S.~Loos, M.~Rabe, C.~Szegedy, and S.~Wilcox.
\newblock Holist: An environment for machine learning of higher order logic theorem proving.
\newblock In \emph{International Conference on Machine Learning}, pages 454--463. PMLR, 2019.

\bibitem[Bibel(2013)]{bibel2013automated}
W.~Bibel.
\newblock \emph{Automated theorem proving}.
\newblock Springer Science \& Business Media, 2013.

\bibitem[Crouse et~al.(2021)Crouse, Abdelaziz, Makni, Whitehead, Cornelio, Kapanipathi, Srinivas, Thost, Witbrock, and Fokoue]{crouse2021deep}
M.~Crouse, I.~Abdelaziz, B.~Makni, S.~Whitehead, C.~Cornelio, P.~Kapanipathi, K.~Srinivas, V.~Thost, M.~Witbrock, and A.~Fokoue.
\newblock A deep reinforcement learning approach to first-order logic theorem proving.
\newblock In \emph{Proceedings of the AAAI Conference on Artificial Intelligence}, volume~35, pages 6279--6287, 2021.

\bibitem[De~Moura et~al.(2015)De~Moura, Kong, Avigad, Van~Doorn, and von Raumer]{de2015lean}
L.~De~Moura, S.~Kong, J.~Avigad, F.~Van~Doorn, and J.~von Raumer.
\newblock The lean theorem prover (system description).
\newblock In \emph{Automated Deduction-CADE-25: 25th International Conference on Automated Deduction, Berlin, Germany, August 1-7, 2015, Proceedings 25}, pages 378--388. Springer, 2015.

\bibitem[First et~al.(2023)First, Rabe, Ringer, and Brun]{first2023baldur}
E.~First, M.~N. Rabe, T.~Ringer, and Y.~Brun.
\newblock Baldur: Whole-proof generation and repair with large language models, 2023.

\bibitem[Han et~al.(2021)Han, Rute, Wu, Ayers, and Polu]{han2021proof}
J.~M. Han, J.~Rute, Y.~Wu, E.~W. Ayers, and S.~Polu.
\newblock Proof artifact co-training for theorem proving with language models.
\newblock \emph{arXiv preprint arXiv:2102.06203}, 2021.

\bibitem[Huang et~al.(2024)Huang, Lin, Liu, Cao, Xin, Wang, Li, Song, and Liang]{huang2024mustard}
Y.~Huang, X.~Lin, Z.~Liu, Q.~Cao, H.~Xin, H.~Wang, Z.~Li, L.~Song, and X.~Liang.
\newblock Mustard: Mastering uniform synthesis of theorem and proof data.
\newblock \emph{arXiv preprint arXiv:2402.08957}, 2024.

\bibitem[Jiang et~al.(2021)Jiang, Li, Han, and Wu]{jiang2021lisa}
A.~Q. Jiang, W.~Li, J.~M. Han, and Y.~Wu.
\newblock Lisa: Language models of isabelle proofs.
\newblock In \emph{6th Conference on Artificial Intelligence and Theorem Proving}, pages 378--392, 2021.

\bibitem[Jiang et~al.(2022{\natexlab{a}})Jiang, Li, Tworkowski, Czechowski, Odrzyg{\'o}{\'z}d{\'z}, Mi{\l}o{\'s}, Wu, and Jamnik]{jiang2022thor}
A.~Q. Jiang, W.~Li, S.~Tworkowski, K.~Czechowski, T.~Odrzyg{\'o}{\'z}d{\'z}, P.~Mi{\l}o{\'s}, Y.~Wu, and M.~Jamnik.
\newblock Thor: Wielding hammers to integrate language models and automated theorem provers.
\newblock \emph{Advances in Neural Information Processing Systems}, 35:\penalty0 8360--8373, 2022{\natexlab{a}}.

\bibitem[Jiang et~al.(2022{\natexlab{b}})Jiang, Welleck, Zhou, Li, Liu, Jamnik, Lacroix, Wu, and Lample]{jiang2022draft}
A.~Q. Jiang, S.~Welleck, J.~P. Zhou, W.~Li, J.~Liu, M.~Jamnik, T.~Lacroix, Y.~Wu, and G.~Lample.
\newblock Draft, sketch, and prove: Guiding formal theorem provers with informal proofs.
\newblock \emph{arXiv preprint arXiv:2210.12283}, 2022{\natexlab{b}}.

\bibitem[Jiang et~al.(2023)Jiang, Li, and Jamnik]{jiang2023multilingual}
A.~Q. Jiang, W.~Li, and M.~Jamnik.
\newblock Multilingual mathematical autoformalization.
\newblock \emph{arXiv preprint arXiv:2311.03755}, 2023.

\bibitem[Kaliszyk et~al.(2018)Kaliszyk, Urban, Michalewski, and Ol{\v{s}}{\'a}k]{kaliszyk2018reinforcement}
C.~Kaliszyk, J.~Urban, H.~Michalewski, and M.~Ol{\v{s}}{\'a}k.
\newblock Reinforcement learning of theorem proving.
\newblock \emph{Advances in Neural Information Processing Systems}, 31, 2018.

\bibitem[Kov{\'a}cs and Voronkov(2013)]{kovacs2013first}
L.~Kov{\'a}cs and A.~Voronkov.
\newblock First-order theorem proving and vampire.
\newblock In \emph{International Conference on Computer Aided Verification}, pages 1--35. Springer, 2013.

\bibitem[Lample et~al.(2022)Lample, Lacroix, Lachaux, Rodriguez, Hayat, Lavril, Ebner, and Martinet]{lample2022hypertree}
G.~Lample, T.~Lacroix, M.-A. Lachaux, A.~Rodriguez, A.~Hayat, T.~Lavril, G.~Ebner, and X.~Martinet.
\newblock Hypertree proof search for neural theorem proving.
\newblock \emph{Advances in neural information processing systems}, 35:\penalty0 26337--26349, 2022.

\bibitem[Liu et~al.(2023)Liu, Shen, Xin, Liu, Yuan, Wang, Ju, Zheng, Yin, Li, et~al.]{liu2023fimo}
C.~Liu, J.~Shen, H.~Xin, Z.~Liu, Y.~Yuan, H.~Wang, W.~Ju, C.~Zheng, Y.~Yin, L.~Li, et~al.
\newblock Fimo: A challenge formal dataset for automated theorem proving.
\newblock \emph{arXiv preprint arXiv:2309.04295}, 2023.

\bibitem[Loos et~al.(2017)Loos, Irving, Szegedy, and Kaliszyk]{loos2017deep}
S.~Loos, G.~Irving, C.~Szegedy, and C.~Kaliszyk.
\newblock Deep network guided proof search.
\newblock \emph{arXiv preprint arXiv:1701.06972}, 2017.

\bibitem[Moura and Ullrich(2021)]{moura2021lean}
L.~d. Moura and S.~Ullrich.
\newblock The lean 4 theorem prover and programming language.
\newblock In \emph{Automated Deduction--CADE 28: 28th International Conference on Automated Deduction, Virtual Event, July 12--15, 2021, Proceedings 28}, pages 625--635. Springer, 2021.

\bibitem[Paulson(1994)]{paulson_isabelle_1994}
L.~C. Paulson.
\newblock \emph{Isabelle a {Generic} {Theorem} {Prover}}.
\newblock Springer Verlag, 1994.

\bibitem[Polu and Sutskever(2020)]{polu2020generative}
S.~Polu and I.~Sutskever.
\newblock Generative language modeling for automated theorem proving.
\newblock \emph{arXiv preprint arXiv:2009.03393}, 2020.

\bibitem[Polu et~al.(2022)Polu, Han, Zheng, Baksys, Babuschkin, and Sutskever]{polu2022formal}
S.~Polu, J.~M. Han, K.~Zheng, M.~Baksys, I.~Babuschkin, and I.~Sutskever.
\newblock Formal mathematics statement curriculum learning.
\newblock \emph{arXiv preprint arXiv:2202.01344}, 2022.

\bibitem[Radford et~al.(2019)Radford, Wu, Child, Luan, Amodei, Sutskever, et~al.]{radford2019language}
A.~Radford, J.~Wu, R.~Child, D.~Luan, D.~Amodei, I.~Sutskever, et~al.
\newblock Language models are unsupervised multitask learners.
\newblock \emph{OpenAI blog}, 1\penalty0 (8):\penalty0 9, 2019.

\bibitem[Schulz(2002)]{schulz2002brainiac}
S.~Schulz.
\newblock E--a brainiac theorem prover.
\newblock \emph{Ai Communications}, 15\penalty0 (2-3):\penalty0 111--126, 2002.

\bibitem[Shao et~al.(2024)Shao, Wang, Zhu, Xu, Song, Zhang, Li, Wu, and Guo]{shao2024deepseekmath}
Z.~Shao, P.~Wang, Q.~Zhu, R.~Xu, J.~Song, M.~Zhang, Y.~Li, Y.~Wu, and D.~Guo.
\newblock Deepseekmath: Pushing the limits of mathematical reasoning in open language models.
\newblock \emph{arXiv preprint arXiv:2402.03300}, 2024.

\bibitem[Shulman(2024)]{Shulman}
M.~Shulman.
\newblock Strange new universes: Proof assistants and synthetic foundations, 2024.

\bibitem[Thakur et~al.(2023)Thakur, Wen, and Chaudhuri]{thakur2023language}
A.~Thakur, Y.~Wen, and S.~Chaudhuri.
\newblock A language-agent approach to formal theorem-proving.
\newblock \emph{arXiv preprint arXiv:2310.04353}, 2023.

\bibitem[{The Coq Development Team}()]{coq}
{The Coq Development Team}.
\newblock Coq.
\newblock URL \url{https://coq.inria.fr}.

\bibitem[Vaswani et~al.(2017)Vaswani, Shazeer, Parmar, Uszkoreit, Jones, Gomez, Kaiser, and Polosukhin]{vaswani2017attention}
A.~Vaswani, N.~Shazeer, N.~Parmar, J.~Uszkoreit, L.~Jones, A.~N. Gomez, {\L}.~Kaiser, and I.~Polosukhin.
\newblock Attention is all you need.
\newblock \emph{Advances in neural information processing systems}, 30, 2017.

\bibitem[Wang and Deng(2020)]{wang2020learning}
M.~Wang and J.~Deng.
\newblock Learning to prove theorems by learning to generate theorems.
\newblock \emph{Advances in Neural Information Processing Systems}, 33:\penalty0 18146--18157, 2020.

\bibitem[Wu et~al.(2021)Wu, Norrish, Walder, and Dezfouli]{wu2021tacticzero}
M.~Wu, M.~Norrish, C.~Walder, and A.~Dezfouli.
\newblock Tacticzero: Learning to prove theorems from scratch with deep reinforcement learning.
\newblock \emph{Advances in Neural Information Processing Systems}, 34:\penalty0 9330--9342, 2021.

\bibitem[Wu et~al.(2020)Wu, Jiang, Ba, and Grosse]{wu2020int}
Y.~Wu, A.~Q. Jiang, J.~Ba, and R.~Grosse.
\newblock Int: An inequality benchmark for evaluating generalization in theorem proving.
\newblock \emph{arXiv preprint arXiv:2007.02924}, 2020.

\bibitem[Wu et~al.(2022)Wu, Jiang, Li, Rabe, Staats, Jamnik, and Szegedy]{wu2022autoformalization}
Y.~Wu, A.~Q. Jiang, W.~Li, M.~Rabe, C.~Staats, M.~Jamnik, and C.~Szegedy.
\newblock Autoformalization with large language models.
\newblock \emph{Advances in Neural Information Processing Systems}, 35:\penalty0 32353--32368, 2022.

\bibitem[Xin et~al.(2023)Xin, Wang, Zheng, Li, Liu, Cao, Huang, Xiong, Shi, Xie, et~al.]{xin2023lego}
H.~Xin, H.~Wang, C.~Zheng, L.~Li, Z.~Liu, Q.~Cao, Y.~Huang, J.~Xiong, H.~Shi, E.~Xie, et~al.
\newblock Lego-prover: Neural theorem proving with growing libraries.
\newblock \emph{arXiv preprint arXiv:2310.00656}, 2023.

\bibitem[Xiong et~al.(2023)Xiong, Shen, Yuan, Wang, Yin, Liu, Li, Guo, Cao, Huang, et~al.]{xiong2023trigo}
J.~Xiong, J.~Shen, Y.~Yuan, H.~Wang, Y.~Yin, Z.~Liu, L.~Li, Z.~Guo, Q.~Cao, Y.~Huang, et~al.
\newblock Trigo: Benchmarking formal mathematical proof reduction for generative language models.
\newblock \emph{arXiv preprint arXiv:2310.10180}, 2023.

\bibitem[Yang et~al.(2024)Yang, Swope, Gu, Chalamala, Song, Yu, Godil, Prenger, and Anandkumar]{yang2024leandojo}
K.~Yang, A.~Swope, A.~Gu, R.~Chalamala, P.~Song, S.~Yu, S.~Godil, R.~J. Prenger, and A.~Anandkumar.
\newblock Leandojo: Theorem proving with retrieval-augmented language models.
\newblock \emph{Advances in Neural Information Processing Systems}, 36, 2024.

\bibitem[Zhao et~al.(2023)Zhao, Li, and Kong]{zhao2023decomposing}
X.~Zhao, W.~Li, and L.~Kong.
\newblock Decomposing the enigma: Subgoal-based demonstration learning for formal theorem proving.
\newblock \emph{arXiv preprint arXiv:2305.16366}, 2023.

\bibitem[Zheng et~al.(2021)Zheng, Han, and Polu]{zheng2021minif2f}
K.~Zheng, J.~M. Han, and S.~Polu.
\newblock Minif2f: a cross-system benchmark for formal olympiad-level mathematics.
\newblock \emph{arXiv preprint arXiv:2109.00110}, 2021.

\end{thebibliography}
